\documentclass{article}

\usepackage{neurips, times} 
\usepackage[numbers]{natbib}
\usepackage{amssymb}

\nipsfinalcopy 

\usepackage{graphicx}
\usepackage{hyperref}
\usepackage{amsmath}
\usepackage{enumitem}
\usepackage{url}

\title{Vibe Coding an LLM-powered Theorem Prover}

\author{%
  Zhe Hou \\
  Griffith University\\
  \texttt{z.hou@griffith.edu.au}
}

\begin{document}

\maketitle

\begin{abstract}
We present Isabellm, an LLM-powered theorem prover for Isabelle/HOL that performs fully automatic proof synthesis. Isabellm works with any local LLM on Ollama and APIs such as Gemini CLI, and it is designed to run on consumer grade computers. The system combines a stepwise prover, which uses large language models to propose proof commands validated by Isabelle in a bounded search loop, with a higher-level proof planner that generates structured Isar outlines and attempts to fill and repair remaining gaps. The framework includes beam search for tactics, tactics reranker ML and RL models, premise selection with small transformer models, micro-RAG for Isar proofs built from AFP, and counter-example guided proof repair. All the code is implemented by GPT 4.1 - 5.2, Gemini 3 Pro, and Claude 4.5. Empirically, Isabellm can prove certain lemmas that defeat Isabelle’s standard automation, including Sledgehammer, demonstrating the practical value of LLM-guided proof search. At the same time, we find that even state-of-the-art LLMs, such as GPT 5.2 Extended Thinking and Gemini 3 Pro struggle to reliably implement the intended fill-and-repair mechanisms with complex algorithmic designs, highlighting fundamental challenges in LLM code generation and reasoning. The code of Isabellm is available at \url{https://github.com/zhehou/llm-isabelle}.
\end{abstract}

\section{Introduction}
Automated reasoning has long been a central goal of computer science, spanning automated theorem proving, model checking, symbolic execution, and satisfiability modulo theories. Classical reasoning engines have achieved remarkable success in well-scoped domains, such as propositional and first-order logic, where decades of algorithmic advances have led to highly optimised SAT and SMT solvers. However, as the expressiveness of the underlying logic increases, full automation becomes significantly harder. Higher-order logics, rich type systems, and large mathematical libraries introduce vast search spaces, subtle dependencies, and non-local reasoning patterns that are difficult to capture with purely symbolic methods. As a result, many practical reasoning systems adopt a hybrid stance, combining automation with human guidance to manage complexity while preserving soundness.

Interactive theorem provers such as Isabelle/HOL~\cite{nipkow2002isabelle} exemplify this trade-off. They provide exceptionally strong correctness guarantees: a proof is accepted only if it type-checks against a small trusted kernel, yielding artefacts far more robust than conventional testing or informal argumentation. The cost of this rigor is well known. Proof development is labour-intensive, requires specialised expertise, and often demands careful orchestration of proof methods and auxiliary lemmas, even in the presence of powerful automation. Isabelle’s ecosystem has substantially mitigated this burden through tools such as Sledgehammer, which combines premise selection, external automated provers, and proof reconstruction to discharge many routine goals \cite{Blanchette2011Sledgehammer,blanchette2010nitpick}. Complementary tools such as Nitpick further assist users by quickly refuting false conjectures or exposing missing assumptions. Nevertheless, fully automatic “push-button” proving for arbitrary higher-order goals remains out of reach: the search space is combinatorially large, the relevant context may involve thousands of facts, and the correct sequence of proof methods is highly problem-dependent.

Recent advances in large language models (LLMs) have reopened this challenge from a different angle. Rather than relying solely on symbolic heuristics, LLMs act as powerful conditional generators that can propose plausible proof steps, tactics, or even full proof scripts based on patterns learned from large corpora. This capability has given rise to a new generation of neuro-symbolic theorem provers that tightly couple generative models with formal verification. In these systems, the language model explores the proof space by proposing candidates, while the proof assistant provides exact, executable feedback by accepting, rejecting, or partially validating those proposals. Early work demonstrated the viability of this approach in settings such as Metamath~\cite{yu2023metamath}, where generative models were combined with guided search and verification loops. Subsequent systems introduced more structured search and learning mechanisms, including HyperTree Proof Search (HTPS), which frames theorem proving as iterative improvement over a dynamically expanding search tree \cite{lample2022hypertree}.

Within the Isabelle ecosystem, this line of work has been enabled by programmatic interfaces and benchmarks that support large-scale interaction and learning, most notably LISA and the PISA protocol for incremental proof execution and data extraction \cite{Jiang2021LISA}. Building on these foundations, recent Isabelle-oriented systems have explored whole-proof generation, verifier-guided repair, and deeper integration with ATP tools, as exemplified by Baldur and Thor \cite{First2023Baldur,Jiang2022Thor}. In parallel, work in other proof assistants, such as LeanDojo for Lean, has highlighted the critical role of retrieval and premise selection when operating over large libraries, as well as the importance of standardised benchmarks such as miniF2F for comparative evaluation \cite{Yang2023LeanDojo,Zheng2021miniF2F}.

This project explores the question
%\begin{quote}
\emph{Can LLM code a theorem prover powered by LLM?}
%\end{quote}
This ``cyclic'' question examines the latest LLMs' capabilities in two aspects: Can they be used to code complex algorithms like a theorem prover? And can they be used to generate formal proofs as a part of the theorem prover? The end goal is fully automatic proof synthesis: given a goal statement (and imports), the system should output a complete, checkable Isabelle proof with no user interaction. Moreover, we aim to make the framework \textbf{laptop-friendly} --- all the current code has been tested on a 2021 Macbook Pro with M1 Pro processor and 32GB RAM. We hope to build an LLM prover that can run on consumer level computers for an average Isabelle/HOL user, rather than something that requires expensive GPU servers. Concretely, the repository implements Isabellm, a modular pipeline with two cooperating layers. The first layer is a stepwise prover that treats proving as sequential decision making: from a proof state, it proposes candidate tactics or proof methods, checks them in Isabelle, and uses search (e.g., beam-style exploration) to reach a terminal solved state. This layer integrates classic Isabelle automation (including Sledgehammer, SMT/ATP backends, and counterexample checking) with learned components such as tactic reranking and premise selection, enabling the prover to scale beyond what raw prompting can hold in context \cite{Blanchette2011Sledgehammer}. The second layer is a planner that attempts to generate structured Isar-style proof outlines and then discharge the resulting subgoals by calling the stepwise prover. This planning layer is motivated by evidence that decomposition and sketching can mitigate myopic step-by-step search, especially for longer proofs, aligning with recent “sketch-first” and recursive proving themes in the literature \cite{Jiang2022Thor,Wang2024POETRY}. The overall design is intended to support rigorous evaluation (benchmarking and regression), reproducibility (scripted proof checking), and continual improvement via logs and learned models.

The remainder of this report describes the system architecture, interfaces, and learning components, and positions Isabellm relative to prior neuro-symbolic provers. We emphasise the technical choices needed to make fully automatic proof generation practical in Isabelle: reliable proof-state interaction, fast verification, premise management at library scale, search control under tight time budgets, and training signals derived from verifier outcomes. We also highlight current limitations, especially around planning and proof repair, and motivate directions for turning verifier feedback into systematic improvements rather than ad hoc prompt tuning.

\section{An LLM-powered Stepwise Prover}

The stepwise prover is the core automated proving engine of the system. Its purpose is to synthesize a complete Isabelle/HOL proof script for a given goal by iteratively proposing proof commands, validating them with Isabelle, and performing bounded search over the resulting proof states. It targets small step proofs, basically playing the same role as Sledgehammer.

\subsection{Preparation}

Proof synthesis is formulated as bounded search in the space of Isabelle proof scripts. A proof script is represented as an ordered sequence
\[
S = \langle s_0, s_1, \dots, s_k \rangle,
\]
where $s_0$ is a lemma declaration of the form \texttt{lemma "<goal>"} and each $s_i$ for $i \ge 1$ is an Isabelle proof command, typically an \texttt{apply}-style tactic.

At each iteration, the prover maintains a beam of candidate proof states of fixed width $B$. Each beam element is a tuple
\[
(\sigma, S, h, n),
\]
where:
\begin{itemize}
  \item $S$ is the current accepted proof prefix,
  \item $\sigma$ is a scalar score used for ranking beam entries,
  \item $h$ is a textual hint representing the current proof state (obtained via \texttt{print\_state}),
  \item $n$ is the number of remaining subgoals reported by Isabelle.
\end{itemize}

The primary score is the number of remaining subgoals $n$. States with fewer subgoals are preferred, reflecting the heuristic that local progress correlates with global proof completion. When $n$ cannot be extracted, a large sentinel value is used.

The search proceeds for at most $D$ expansion rounds (maximum depth), or until a proof is completed, or until a global wall-clock timeout is reached.

%\subsection{Theory construction and Isabelle execution}

Each candidate proof command is validated by constructing a temporary Isabelle theory and invoking Isabelle through a persistent server connection. For a proof prefix
\[
S = \langle s_0, s_1, \dots, s_k \rangle
\]
and a candidate command $c$, the prover constructs a theory of the form
\[
\texttt{theory Scratch imports Main begin}
\]
\[
\quad s_0
\]
\[
\quad s_1
\]
\[
\quad \vdots
\]
\[
\quad s_k
\]
\[
\quad c
\]
% \[
% \texttt{end}
% \]
followed by either:
\begin{itemize}
  \item \texttt{print\_state} and \texttt{sorry} for intermediate steps, or
  \item no trailing command for proof finishers.
\end{itemize}

The theory is executed using Isabelle’s \texttt{use\_theories} command. A candidate step is considered successful if Isabelle accepts the theory without error. For intermediate steps, the prover extracts the printed proof state and parses the number of remaining subgoals.

To avoid redundant exploration, proof states are deduplicated using a fingerprint computed as
\[
\mathsf{fp}(h) = \textsf{SHA1}(\textsf{normalize}(h)),
\]
where normalization removes irrelevant whitespace. Only one beam entry per fingerprint is retained.

\subsection{Candidate generation via language models}

Candidate proof commands are generated by large language models (LLMs) acting as conditional proposal mechanisms inside the stepwise search loop. The LLMs are not used to generate complete proofs in one shot; instead, they are queried repeatedly to propose small, locally valid proof commands that are then checked by Isabelle. This design ensures that all correctness guarantees are enforced by the proof assistant rather than the language model.

\paragraph{Invocation model.}
LLMs are invoked through a unified abstraction layer that supports both local and remote models (e.g., via HTTP or CLI interfaces). Each invocation specifies:
\begin{itemize}
  \item a system prompt defining the allowed action grammar,
  \item a user prompt encoding the current proof context,
  \item decoding parameters including temperature, maximum tokens, and number of samples.
\end{itemize}

The prover does not assume any particular model architecture. Instead, it treats the LLM as a black-box conditional generator that maps a textual proof context to a short list of candidate proof commands.

\paragraph{Prompting modes.}
Two distinct prompting modes are used, corresponding to different phases of proof search:

\begin{enumerate}
  \item \textbf{Step mode}: generates intermediate proof commands of the form
  \[
  \texttt{apply <tactic>}.
  \]
  These commands are intended to reduce or transform the current set of subgoals without closing the proof.

  \item \textbf{Finish mode}: generates proof-closing commands such as
  \[
  \texttt{by simp}, \quad \texttt{by auto}, \quad \texttt{by (metis ...)}, \quad \texttt{done}.
  \]
  Finish mode is invoked when the prover believes the remaining subgoals may be solvable in a single step.
\end{enumerate}

Each mode uses a distinct system prompt to tightly constrain the syntactic form of the output.

\paragraph{System prompts.}
The system prompt explicitly instructs the model to:
\begin{itemize}
  \item output multiple candidate commands (typically between 3 and 8),
  \item output one command per line,
  \item restrict output to a fixed syntactic grammar (\texttt{apply} commands in step mode, \texttt{by}/\texttt{done} in finish mode),
  \item avoid explanations, comments, or natural language text.
\end{itemize}

By enforcing these constraints at the prompt level, the prover reduces the burden on downstream parsing and significantly lowers the rate of invalid Isabelle calls.

\paragraph{User prompt construction.}
The user prompt encodes the current proof context as a structured text block containing:

\begin{enumerate}
  \item \textbf{Goal statement:} the original lemma to be proved.
  \item \textbf{Accepted proof steps:} the prefix of proof commands that have already been validated by Isabelle.
  \item \textbf{Current proof state:} the latest \texttt{print\_state} output, showing remaining subgoals.
  \item \textbf{Helpful facts:} a curated list of lemma names obtained from premise mining and retrieval.
\end{enumerate}

This information is presented in a consistent, machine-readable format to stabilize the model’s behavior across different proof states. The proof state is included verbatim to expose the logical structure of remaining goals, while the helpful facts provide soft guidance without enforcing hard constraints.

\paragraph{Decoding control and exploration.}
LLM decoding is stochastic and controlled primarily via temperature. To balance exploration and exploitation, the prover adapts the temperature dynamically based on search stagnation.

Let $s$ denote the number of consecutive search depths without improvement in the minimum number of subgoals. The decoding temperatures are adjusted as:
\[
T_{\text{step}} = \min(0.9,\ 0.5 + 0.1s), \qquad
T_{\text{finish}} = \min(0.6,\ 0.2 + 0.05s).
\]

At low stagnation, temperatures are kept low to favor deterministic, high-probability commands. As stagnation increases, temperatures rise to encourage syntactic and semantic diversity in proposed commands.

\paragraph{Candidate sampling.}
For each invocation, the model is asked to generate a small batch of candidates in a single call. This is more efficient than repeated single-sample calls and allows the prover to exploit intra-batch diversity. The number of candidates per call is fixed and independent of beam width, ensuring predictable computational cost.

\paragraph{Post-processing and sanitisation.}
The raw output of the language model is subjected to strict post-processing:

\begin{itemize}
  \item code blocks, numbering, bullets, and extraneous whitespace are removed;
  \item only lines beginning with approved prefixes are retained;
  \item commands exceeding a fixed character limit are discarded;
  \item duplicate commands are removed while preserving order.
\end{itemize}

Any output that does not conform to the expected grammar is silently ignored. This conservative filtering ensures that only syntactically plausible Isabelle commands reach the expensive validation stage.

\paragraph{Integration with reranking.}
If a learned reranker is available, each candidate command is immediately featurised and scored before Isabelle evaluation. The reranker score is used to reorder candidates, so that more promising commands are checked first. This interaction allows learning to influence the search without bypassing symbolic validation.

\paragraph{Heuristic variant injection.}
When stagnation exceeds a threshold, the prover augments LLM-generated candidates with heuristic templates such as:
\[
\texttt{apply (induction xs)}, \quad \texttt{apply (cases x)}, \quad \texttt{apply (rule some\_lemma)}.
\]
These templates are instantiated using variable names extracted from the current proof state. This mechanism acts as a lightweight fallback that injects domain-relevant structure even when the LLM fails to propose it explicitly.

\paragraph{Design rationale.}
The candidate generation mechanism deliberately restricts the expressive power of the language model. By limiting outputs to short, grammar-constrained commands and validating every proposal symbolically, the system transforms the LLM from an unreliable proof generator into a probabilistic proposal distribution over local proof actions. This design keeps the overall prover sound, debuggable, and amenable to learning-based improvement.

\subsection{Beam expansion and scoring}

For each beam state, the prover generates a finite set of candidate next steps. Each candidate is independently validated with Isabelle. Successful candidates yield new beam entries with updated proof prefixes and proof states.

Let $\mathcal{C}$ be the multiset of all successful expansions from the current beam. The next beam is constructed by sorting $\mathcal{C}$ lexicographically by:
\[
(n,\ |S|),
\]
where $n$ is the number of remaining subgoals and $|S|$ is the length of the proof script. The best $B$ entries with distinct fingerprints are retained.

This yields a best-first search over proof states, constrained by beam width and depth.

Monte Carlo tree search is tested by not used because it is much slower than beam search, and as alluded above, the goal of the stepwise prover is to prove small step goals rather than complex goals.

\subsection{Learning-based tactic reranking}

The stepwise prover employs a learning-based reranker to bias the ordering of candidate proof commands proposed by the language model. The reranker operates as a lightweight scoring function that predicts the likelihood that a candidate command will make progress when applied to the current proof state. Importantly, the reranker does \emph{not} replace symbolic validation: all candidates are still checked by Isabelle. Its role is purely to reduce the effective branching factor of the search.

\paragraph{Reranking interface.}
At runtime, the reranker is exposed as a function
\[
f_\theta : \mathbb{R}^d \rightarrow [0,1],
\]
where $d$ is the feature dimension and $f_\theta(x)$ estimates the probability that a candidate step will succeed. Higher scores indicate higher priority during beam expansion. The reranker score is incorporated into candidate ordering by adjusting the heuristic score:
\[
\mathrm{score}(c) = \mathrm{heuristic}(c) - \lambda \cdot f_\theta(x_c),
\]
where $\lambda > 0$ is a fixed weight (configured via environment variables) and $x_c$ is the feature vector extracted for candidate $c$.

\paragraph{Feature representation.}
Each candidate step is represented by a fixed-length numeric feature vector composed of four groups:

\begin{enumerate}
  \item \textbf{Search context features:}
  \[
  (\text{depth},\ n_{\text{sub}},\ t_{\text{elapsed}},\ \text{cache\_hit}),
  \]
  where depth is the current search depth, $n_{\text{sub}}$ is the number of remaining subgoals (if known), and cache\_hit indicates reuse of cached Isabelle results.

  \item \textbf{Goal and state flags:}
  Binary indicators extracted from the goal and proof state, including:
  \[
  \{\text{is\_listy},\ \text{is\_natty},\ \text{is\_sety},\ \text{has\_quantifier},\ \text{is\_boolean}\}.
  \]
  These capture coarse structural properties of the goal.

  \item \textbf{Tactic prefix encoding:}
  A one-hot encoding over a fixed vocabulary of tactic prefixes (e.g., \texttt{apply simp}, \texttt{apply auto}, \texttt{apply (induction)}, \texttt{apply (cases)}, \texttt{apply (rule)}, etc.). This represents the syntactic class of the proposed command.

  \item \textbf{Premise interaction features:}
  Numeric summaries derived from premise selection (described later), including cosine similarity statistics and overlap between candidate-referenced lemmas and retrieved premises.
\end{enumerate}

The final feature vector is padded or truncated to a fixed dimension expected by the trained model. This makes the reranker robust to incremental feature extensions.

\paragraph{Training data.}
Training data for the reranker is collected automatically during proof search. Every attempted candidate command generates a labeled example
\[
(x_c, y_c),
\]
where $x_c$ is the feature vector described above and
\[
y_c =
\begin{cases}
1 & \text{if Isabelle accepts the step}, \\
0 & \text{otherwise}.
\end{cases}
\]
Both successful and failed attempts are logged. This produces a highly imbalanced but extremely large dataset reflecting real search behavior rather than curated proofs.

\paragraph{Supervised learning.}
The system supports multiple supervised learning backends:

\begin{itemize}
  \item \textbf{Logistic regression:} a linear baseline for fast iteration and interpretability.
  \item \textbf{Gradient-boosted trees (XGBoost):} used to model non-linear interactions between features.
\end{itemize}

For probabilistic classifiers, the predicted probability $P(y=1 \mid x)$ is used directly as $f_\theta(x)$. For regressors, outputs are normalized to $[0,1]$.

\paragraph{Offline reinforcement learning.}
Beyond supervised learning, the system supports offline reinforcement learning from logged trajectories. In this formulation, proof search is treated as a Markov decision process:
\[
(s_t, a_t, r_t, s_{t+1}),
\]
where $s_t$ is the proof state, $a_t$ is a candidate tactic, and the reward $r_t$ is derived from subgoal reduction or proof completion.

Two algorithms are implemented:

\begin{enumerate}
  \item \textbf{Advantage-Weighted Regression (AWR):} candidates are weighted by estimated advantage, favoring steps that lead to faster subgoal reduction.
  \item \textbf{Fitted Q-learning (DQN-style):} learns an action-value function $Q(s,a)$ from logged transitions and uses it as a reranking signal.
\end{enumerate}

The learned models are exported either as TorchScript modules or joblib artifacts and loaded dynamically at runtime.

\paragraph{Design rationale.}
The reranker is intentionally shallow and fast. Its purpose is not to prove theorems independently, but to encode empirical regularities of what tends to work in Isabelle, allowing the symbolic prover to focus its computational budget on promising branches.

\subsection{Premise selection with neural encoders}

Premise selection is used to supply the prover and the language model with a small, relevant subset of lemmas from the surrounding theory and library. Rather than relying solely on Isabelle’s internal heuristics, the system implements a retrieval-based premise selector with optional neural encoders.

\paragraph{Premise index.}
All candidate premises are stored in an in-memory index as pairs
\[
(\ell_i, t_i),
\]
where $\ell_i$ is a lemma identifier and $t_i$ is its textual representation. Each premise may also carry metadata such as source file and local context.

\paragraph{Two-stage retrieval.}
Premise selection proceeds in two stages:

\begin{enumerate}
  \item \textbf{Select stage:} a fast, recall-oriented retrieval that produces a pool of $K_{\text{select}}$ candidate premises.
  \item \textbf{Rerank stage:} an optional precision-oriented rescoring of the top $K_{\text{rerank}}$ premises.
\end{enumerate}

\paragraph{Select stage encoders.}
The select stage supports three backends:

\begin{itemize}
  \item \textbf{TF--IDF cosine similarity:} when scikit-learn is available, premises are vectorized using TF--IDF and scored by cosine similarity.
  \item \textbf{Token overlap (fallback):} when no external libraries are available, Jaccard overlap over token sets is used.
  \item \textbf{Neural bi-encoder:} when a trained encoder is present, both goals and premises are embedded into a shared vector space using a sentence-transformer model. Cosine similarity is computed as
  \[
  \mathrm{sim}(g, p) = \frac{\langle e(g), e(p) \rangle}{\|e(g)\|\|e(p)\|}.
  \]
\end{itemize}

All embeddings are computed once during index finalization and cached in memory.

\paragraph{Rerank stage.}
The rerank stage optionally applies a cross-encoder that scores pairs $(g, p)$ jointly. Given a batch of premise candidates $\{p_i\}$ and a goal $g$, the cross-encoder computes scores
\[
r_i = \mathrm{CE}(g, p_i),
\]
which are used to reorder premises. If no reranker is available, the select score is reused.

\paragraph{Training premise encoders.}
Training data for premise selection is extracted from successful proof attempts. For a given goal $g$:

\begin{itemize}
  \item Positive premises are those explicitly referenced in successful proof steps.
  \item Negative premises are sampled from the retrieval pool but not used in the proof.
\end{itemize}

The bi-encoder is trained using contrastive learning (e.g., Multiple Negatives Ranking Loss), encouraging
\[
\mathrm{sim}(g, p^+) > \mathrm{sim}(g, p^-)
\]
for positive premise $p^+$ and negative premise $p^-$. The cross-encoder is trained using supervised regression or classification over $(g, p)$ pairs.

\paragraph{Integration with proof search.}
Selected premises are used in two ways:

\begin{enumerate}
  \item They are injected verbatim into the language model prompt as contextual hints.
  \item Summary statistics (top similarity, mean similarity, overlap with candidate-referenced lemmas) are appended to the reranker feature vector.
\end{enumerate}

This tight coupling allows premise selection and tactic reranking to reinforce each other without hard constraints.

\paragraph{Design rationale.}
Premise selection is treated as a soft guidance mechanism rather than a hard filter. By exposing both the language model and the reranker to retrieved premises, the system benefits from retrieval while remaining robust to retrieval errors, a critical property in large and evolving Isabelle libraries.

\subsection{Post-processing}

Before executing expensive Isabelle checks, candidate steps may be pruned using lightweight refutation tools. Quickcheck and Nitpick are optionally invoked to detect counterexamples. If either tool produces a counterexample, the candidate is discarded without further evaluation.

This pruning is applied conservatively, primarily for goals involving Boolean structure or quantifiers, and at configurable intervals to control overhead.

%\subsection{Caching}

To reduce repeated Isabelle invocations, the prover employs two levels of caching:

\begin{itemize}
  \item a per-run cache keyed by $(S, c)$,
  \item a global bounded cache shared across runs.
\end{itemize}

Cached entries store success flags, subgoal counts, proof state hints, and timing information. Cache hits are logged and later used as features during reranker training.

%\subsection{Proof minimisation and variants}

After a successful proof is found, the prover attempts to simplify it. The minimisation procedure applies the following transformations greedily under a small timeout:

\begin{enumerate}
  \item collapse the proof to a single-line proof if possible,
  \item remove unused facts from \texttt{simp} or \texttt{metis} calls,
  \item delete redundant intermediate steps,
  \item reattempt single-line proofs.
\end{enumerate}

Additionally, the prover attempts to convert unstructured \texttt{apply}-style proofs into structured Isar proofs using simple skeleton templates, improving readability without sacrificing automation.

% \subsection{Continual learning}

% Every candidate attempt is logged in a structured JSON format. Logged fields include the goal, proof prefix, candidate command, success flag, subgoal counts, timing, cache usage, retrieved premises, and reranker scores.

% These logs serve as supervised and offline reinforcement learning data for training both tactic rerankers and premise selection models, closing the loop between proof search and learned guidance.

\section{An LLM-powered Proof Planner}
\label{sec:planner}

This section describes the proof planner, which operates at a higher level of abstraction than the stepwise prover. Whereas the stepwise prover performs bounded search over individual proof commands, the planner reasons over \emph{structured Isar proofs} that explicitly decompose a complex goal into intermediate claims and subproofs. The planner aims to synthesize a proof outline that captures the global structure of the argument and then to systematically eliminate any remaining gaps until a fully verified proof is obtained.

Formally, given a goal formula $G$, the planner seeks a proof script
\[
P = \langle s_0, s_1, \dots, s_m \rangle
\]
such that $s_0$ is a lemma declaration for $G$, $P$ is a well-formed Isar proof, and Isabelle verifies $P$ without unresolved gaps. Unlike the stepwise prover, intermediate scripts may contain placeholders (\texttt{sorry}), which are treated as explicit unknowns to be filled by subsequent phases.

\subsection{High-level planning model}

The planner operates in one of two modes:
\begin{itemize}
  \item \textbf{Outline mode}, which returns a structured Isar proof skeleton that may contain gaps.
  \item \textbf{Auto mode}, which iteratively applies outline generation, gap filling, repair, and regeneration until either a verified proof is produced or a global budget is exhausted.
\end{itemize}

Throughout planning, Isabelle is accessed through a persistent server session. Let $\mathcal{I}$ denote this session. All outline checking, filling, and verification steps are executed against $\mathcal{I}$, amortizing initialization costs and enabling repeated theory construction under tight budgets.

%\subsection{Outline generation as structured hypothesis synthesis}

\paragraph{Diversified outline sampling.}
Outline generation is performed by querying a language model to produce candidate Isar skeletons. Let $\mathcal{T} = \{t_1, \dots, t_r\}$ be a fixed set of sampling temperatures. For each $t_i \in \mathcal{T}$, the planner samples up to $k$ candidate outlines, yielding a multiset
\[
\mathcal{S} = \bigcup_{i=1}^r \mathcal{S}_{t_i},
\]
where $\mathcal{S}_{t_i}$ are the samples at temperature $t_i$. This mechanism approximates sampling from a mixture distribution over proof structures, trading determinism for diversity.

\paragraph{Prompt-level conditioning.}
Each sampling call conditions on:
\begin{enumerate}
  \item the target goal $G$,
  \item an optional set of recommended lemmas $H = \{h_1,\dots,h_m\}$,
  \item a system constraint requiring a single Isar proof block.
\end{enumerate}
Hints are injected \emph{before} generation, allowing them to influence the global structure (e.g.\ choice of induction variable) rather than only local steps.

\paragraph{Normalization into canonical skeletons.}
Raw model outputs are normalized into canonical skeletons
\[
S = (\text{header}, \text{body}, \text{holes}),
\]
where:
\begin{itemize}
  \item the header is forced to match $G$,
  \item the body is a syntactically complete Isar proof,
  \item all missing justifications are replaced by \texttt{sorry},
  \item inline proofs (\texttt{by \dots}) may be rewritten into holes when outline enforcement is enabled.
\end{itemize}

This normalization guarantees that every candidate outline admits a well-defined set of gaps and can be checked by Isabelle without semantic ambiguity.

\paragraph{Hole extraction.}
Let $\mathrm{holes}(S) = \{h_1, \dots, h_\ell\}$ denote the set of maximal \texttt{sorry} spans in $S$. Each hole $h_i$ is treated as an independent subproblem during filling and repair.

The planner exposes the following parameters for experimental control:
\begin{itemize}
  \item outline diversity parameters $(k, \mathcal{T})$,
  \item enforcement of explicit holes,
  \item use of structural templates,
  \item hint sources and retrieval limits,
  \item scoring weights $(\alpha,\beta,\gamma)$,
  \item stage caps $(c_1, c_2)$ and global time budgets.
\end{itemize}

Together, these parameters define a search space over proof structures and repair strategies, enabling systematic evaluation of planning effectiveness.

\subsection{Micro-RAG: lightweight retrieval-augmented guidance for planning}
\label{subsec:micro-rag}

The proof planner incorporates a deliberately lightweight form of retrieval-augmented generation (RAG), referred to as \emph{micro-RAG}, whose purpose is to bias outline generation and repair without introducing heavyweight neural retrieval or large external indices. The design goal is to provide \emph{structural and semantic hints} that are cheap to compute, stable across runs, and suitable for execution on a laptop-scale environment.

Unlike premise selection in the stepwise prover, which is tightly coupled to tactic-level decision making, micro-RAG operates at the level of \emph{proof planning}. Its outputs influence which proof structures are proposed and how gaps are repaired, but never act as hard constraints.

\subsubsection{Hint sources and representation}

Micro-RAG aggregates hints from two independent sources:
\[
H = H_{\text{ctx}} \cup H_{\text{lex}},
\]
where each hint $h \in H$ is a symbolic identifier (typically a lemma name or theorem constant).

\paragraph{Context-derived hints.}
The context-derived hint set $H_{\text{ctx}}$ is extracted directly from Isabelle. Given a goal $G$, the planner constructs a minimal theory that opens the goal and requests a proof state printout. From this state block, it extracts:
\begin{itemize}
  \item locally bound facts,
  \item assumptions,
  \item previously introduced lemmas available in the current context.
\end{itemize}

Let $\mathsf{CtxFacts}(G)$ denote the multiset of symbols appearing in this state. The planner applies syntactic normalization (removing duplicates, stripping qualifiers, filtering trivial facts) and truncates the list to a fixed budget $k_{\text{ctx}}$. The resulting set is
\[
H_{\text{ctx}} := \mathsf{TopK}(\mathsf{CtxFacts}(G), k_{\text{ctx}}).
\]

These hints reflect what Isabelle itself considers locally relevant, making them particularly effective for guiding structural proof choices such as induction variables or case distinctions.

\paragraph{Lexicon-derived hints.}
The lexicon-derived hint set $H_{\text{lex}}$ is obtained from a precomputed \emph{hint lexicon}, stored as a JSON map:
\[
\mathcal{L} : \text{token} \mapsto \{(h, w_h)\},
\]
where each entry associates a token (e.g.\ function name, constructor, type name) with a weighted list of lemma identifiers.

At runtime, the planner tokenizes the goal $G$ into a multiset $\mathrm{Tok}(G)$. For each token $t \in \mathrm{Tok}(G)$ present in $\mathcal{L}$, it retrieves the associated lemma list and accumulates scores:
\[
\mathrm{score}(h) = \sum_{t \in \mathrm{Tok}(G)} w_{t,h}.
\]
The lexicon-derived hint set is then
\[
H_{\text{lex}} := \mathsf{TopK}\big(\{(h,\mathrm{score}(h))\},\ k_{\text{lex}}\big),
\]
where $k_{\text{lex}}$ corresponds to \texttt{hintlex\_top}.

The lexicon itself is mined offline from large Isabelle corpora (e.g.\ AFP) by correlating goal tokens with lemmas appearing in successful Isar proofs. Importantly, this process is entirely symbolic and does not require neural encoders.

\subsubsection{Hint aggregation and normalization}

The combined hint set is formed as:
\[
H := \mathsf{Dedup}\big(H_{\text{ctx}} \cup H_{\text{lex}}\big),
\]
followed by truncation to a global cap $k_{\text{hint}}$. The planner preserves relative ordering by source priority, typically favoring context-derived hints over lexicon hints.

No attempt is made to ensure completeness or optimality of $H$. Micro-RAG is explicitly heuristic and biased toward stability and low variance rather than maximal recall.

\subsubsection{Integration into outline generation}

During outline generation, the hint set $H$ is injected at the \emph{prompt level}. Specifically, the system prompt remains unchanged, but the user prompt includes an explicit preference clause:
\[
\texttt{HINTS: Prefer using } h_1, \dots, h_m \texttt{ if applicable.}
\]

This instruction is advisory rather than mandatory. The language model is free to ignore hints, but empirical behavior shows that such soft conditioning often affects high-level proof choices, such as selecting an induction principle or reusing a characteristic lemma.

Hints are injected \emph{before} outline sampling, allowing them to influence global structure rather than being retrofitted during repair.

\subsubsection{Integration into repair and regeneration}

Micro-RAG is also used during repair and whole-proof regeneration. In these phases, hints serve two roles:

\begin{enumerate}
  \item \textbf{Repair conditioning:} when regenerating a block or subproof, the hint set is included alongside effective goals and error diagnostics, biasing the LLM toward known useful lemmas.
  \item \textbf{Scoring signal:} usage of hint lemmas inside an outline contributes positively to the outline’s composite score via the $\mathrm{hint\_bonus}$ term.
\end{enumerate}

Formally, let $\mathrm{Use}(S,H)$ be the number of distinct hints from $H$ that appear syntactically in skeleton $S$. The bonus term is computed as:
\[
\mathrm{hint\_bonus}(S) = \min\big(\mathrm{Use}(S,H),\ k_{\text{hint}}\big),
\]
ensuring bounded influence.

\subsubsection{Design rationale and limitations}

Micro-RAG deliberately avoids heavyweight retrieval mechanisms such as dense embedding indices or cross-encoders. This choice reflects several considerations:

\begin{itemize}
  \item planning decisions are coarse-grained and benefit more from symbolic cues than from fine-grained semantic similarity;
  \item planner prompts are short and benefit from stable, low-noise hints;
  \item laptop-scale execution precludes large neural indices.
\end{itemize}

As a result, micro-RAG should be viewed as a \emph{structural biasing mechanism} rather than a replacement for premise selection in tactic-level proving. Its strength lies in nudging the planner toward familiar proof schemas while leaving all correctness checking to Isabelle.

\subsection{Gap filling and counterexample-guided proof repair}
\label{subsec:cegis}

This subsection specifies the planner’s fill+repair mechanism in the same verifier-in-the-loop, algorithmic style as the stepwise prover. The key design goal is to treat a partially correct Isar outline as a \emph{structured search object} and to transform each remaining \texttt{sorry} into a sequence of bounded synthesis problems under Isabelle validation.

\subsubsection{Objects and invariants}

Let $P$ denote the current Isar script (a sequence of lines). A \emph{hole} is a maximal span
\[
h = [a,b) \subseteq \{0,\dots,|P|\}
\]
that corresponds to a \texttt{sorry} token occurrence in the text. Let $\mathrm{Holes}(P)=\{h_1,\dots,h_\ell\}$.

The planner maintains the following invariants:
\begin{enumerate}
  \item \textbf{Verifier gate:} a change is \emph{committed} only if the full script verifies in Isabelle.
  \item \textbf{Local progress allowance:} a change that does not verify may still be retained as \emph{partial progress}, but it must be immediately normalized by opening new explicit holes so that subsequent iterations can target them.
  \item \textbf{Stable hole identity:} each hole has a stable identifier used to track repair stage and attempts across textual edits.
\end{enumerate}

\paragraph{Stable hole identifier.}
Since line indices drift after edits, each hole is keyed by a windowed fingerprint:
\[
\mathrm{hid}(P,h) = \mathrm{SHA1}\big(P[\max(0,a-w):\min(|P|,b+w)]\big)[:16],
\]
where $w$ is a fixed character window. This identifier is used by the planner’s state
\[
\Pi = (\mathrm{stage}[\mathrm{hid}],\ \mathrm{tries}[(\mathrm{hid},s)],\ \mathrm{focus}),
\]
where $\mathrm{stage}[\mathrm{hid}] \in \{0,1,2\}$ is the current repair stage for that hole, $\mathrm{tries}$ counts attempts per stage, and $\mathrm{focus}$ optionally pins the traversal to a specific hole after partial progress.

\subsubsection{Effective goal extraction}

For each hole $h$ in script $P$, the planner computes an \emph{effective goal} $G_h$ by querying Isabelle for the proof state immediately preceding $h$.

Concretely, let $P_{\prec h}$ be the prefix of $P$ ending at the location just before \texttt{sorry} in $h$. The planner constructs a temporary theory that executes $P_{\prec h}$ and ends with \texttt{print\_state} (or an equivalent state-printing command) so that Isabelle emits a textual proof state block. Let $\mathsf{State}(P,h)$ denote that block. The effective goal is computed as a text extraction function:
\[
G_h := \mathsf{EffGoal}\big(\mathsf{State}(P,h),\ G,\ P,\ h\big),
\]
which prefers (i) the currently active subgoal statement and (ii) a consistent fallback to the original goal $G$ when state parsing fails.

This transforms global filling into localized subproblems:
\[
G \leadsto \{G_{h_1},\dots,G_{h_\ell}\}.
\]

\subsubsection{Fill: calling the stepwise prover as a local synthesizer}

Given a hole $h$ and its effective goal $G_h$, the planner first attempts a \emph{fill} step by calling the stepwise prover with a small budget and shallow search parameters:
\[
\mathsf{Solve}(G_h;\ \text{budget},\text{depth},\text{beam},\text{facts},\dots) \;\Rightarrow\; \text{a sequence of commands } C.
\]
The returned commands are post-processed to extract:
\begin{itemize}
  \item apply-steps: $A = [c \in C \mid c \text{ begins with \texttt{apply}}]$,
  \item a finisher: $f \in C$ such that $f$ begins with \texttt{by } or equals \texttt{done}.
\end{itemize}

The planner then attempts to splice these commands into the hole region. Two cases are distinguished:

\paragraph{Case 1: finisher available.}
If $f$ is present, the hole is replaced by an indented block consisting of $A$ followed by $f$.
Let $\mathsf{Replace}(P,h,\Delta)$ denote the script obtained by replacing span $h$ with text $\Delta$.
The fill candidate is
\[
P' := \mathsf{Replace}(P,h,\mathrm{Indent}(A \mathbin{+\!\!+} [f])).
\]
If Isabelle verifies $P'$, the fill commits.

\paragraph{Case 2: apply-only progress.}
If only apply-steps are available, the planner \emph{never} declares success immediately. Instead it treats apply-only output as a transformation that may reduce subgoals but does not close the local proof obligation. Moreover, apply-steps are not syntactically admissible everywhere in Isar (notably under \texttt{have}/\texttt{show}/\texttt{obtain} headings in ``prove'' mode). Therefore the planner performs a structural placement check:
\begin{enumerate}
  \item if the hole is in a context where \texttt{apply} is legal, it inserts $A$ and then forces a new explicit gap via \texttt{sorry};
  \item otherwise, it wraps $A$ into a tiny subproof (e.g.\ \texttt{proof -} \ldots \texttt{sorry} \texttt{qed}) so that the script remains well-formed.
\end{enumerate}
In both cases, the result is treated as \emph{partial progress} and fed to the next normalization step below.

\subsubsection{Partial progress normalization: opening minimal sorries}

A central engineering choice is that the planner never continues search on scripts that are ``half-edited'' without explicit holes. Whenever a fill or repair attempt changes the script but fails global verification, the planner calls a normalization operator:
\[
(P^{\star},\ \mathsf{opened}) := \mathsf{OpenMinimalSorries}(P).
\]

Intuitively, $\mathsf{OpenMinimalSorries}$ scans for failing tactic lines (typically \texttt{apply} sequences or \texttt{by} commands) and replaces them with \texttt{sorry} in a manner that preserves the Isar structure:
\begin{itemize}
  \item if an \texttt{apply}-sequence is embedded under a \texttt{have}/\texttt{show} head, it is converted into a local \texttt{proof -} \ldots \texttt{sorry} \texttt{qed} block rather than leaving raw \texttt{apply} lines;
  \item otherwise, the failing line is replaced by a \texttt{sorry} line at matching indentation.
\end{itemize}

If any hole is opened, the planner updates focus to the nearest newly created hole around the old location (to ensure continuity of effort) and resumes from that hole.

\subsubsection{Repair: CEGIS over structured block edits}

If fill fails (or is skipped due to current stage), the planner invokes \texttt{try\_cegis\_repairs}. This procedure implements a bounded, CEGIS-style loop over \emph{block-level edits} rather than single tactics.

\paragraph{Repair search state.}
Let $P$ be the current script and $h$ the target hole. The repair procedure maintains:
\begin{itemize}
  \item $P_t$: current script candidate (initially $P$),
  \item $S_0 := \mathsf{State}(P,h)$: initial state block,
  \item a time budget function $\mathrm{left}()$ derived from $\texttt{repair\_budget\_s}$,
  \item a \emph{prior failure store} $\mathcal{M}$ mapping block type $\to$ a bounded list of previously tried block candidates, used as a ban list for LLM prompting.
\end{itemize}

\paragraph{Retargeting to earliest failure anchor.}
Although repair is parameterized by the hole span, the actual syntactic error often occurs earlier than the \texttt{sorry}. Therefore repair first computes an anchor line:
\[
\ell_{\mathrm{anchor}} := \mathsf{EarliestFailureLine}(P,h),
\]
and sets the repair focus line to a clamped index near $\ell_{\mathrm{anchor}}$. This converts repair from ``edit exactly at the hole'' to ``edit the earliest block that causes the hole to become unprovable.''

\paragraph{Block types and stages.}
Repair considers three block types, aligned with stage escalation:
\begin{enumerate}
  \item \textbf{Stage 1 (have/show micro-block):} find an enclosing \texttt{have}/\texttt{show}/\texttt{obtain} block around the focus line and attempt to regenerate that micro-block.
  \item \textbf{Stage 2a (case block):} if a \texttt{cases}/\texttt{case} structure encloses the focus line, attempt to regenerate the case block.
  \item \textbf{Stage 2b (subproof):} if a \texttt{proof} \ldots \texttt{qed} subproof encloses the focus line, attempt to regenerate the entire subproof.
\end{enumerate}
Whole-proof regeneration is handled by the outer driver as stage 3.

\paragraph{Counterexample hints.}
Before proposing a new block, repair extracts \emph{counterexample-oriented hints}:
\[
\mathrm{CE} := \mathsf{CounterexampleHints}(\mathcal{I}, S_0),
\]
which may include variable bindings and definitional expansions suggested by Quickcheck/Nitpick-style diagnostics. These hints are treated as soft constraints in the LLM prompt.

\paragraph{LLM proposal and strict deduplication.}
For a block region $B$ (a contiguous span of lines $[i,j)$), repair builds a prompt containing:
\begin{itemize}
  \item the effective goal $G_h$,
  \item normalized Isabelle error messages observed on the current script,
  \item the extracted proof context near the block,
  \item the current block text $B$,
  \item counterexample hints $\mathrm{CE}$,
  \item a \emph{prior failed block list} from $\mathcal{M}$.
\end{itemize}

The LLM outputs a candidate replacement block $\widehat{B}$. Repair computes a fingerprint
\[
\mathrm{fp}(\widehat{B}) := \mathrm{SHA1}(\mathrm{normalize}(\widehat{B})),
\]
and rejects $\widehat{B}$ immediately if $\mathrm{fp}(\widehat{B})$ matches any fingerprint already stored in $\mathcal{M}$ for that block type. This ``strict deduplication'' prevents pathological repetition under stochastic decoding.

\paragraph{Wrapper stripping and block canonicalization.}
Since LLMs frequently include extraneous wrappers (e.g.\ emitting a full lemma instead of a sub-block), repair applies type-specific stripping operators:
\[
\widehat{B}' := \mathsf{StripToType}(\widehat{B}, \mathrm{type}),
\]
so that the replacement has the same syntactic granularity as the original block.

\paragraph{Repair verification gate.}
A candidate script is formed by substitution
\[
P' := \mathsf{ReplaceLines}(P, [i,j), \widehat{B}').
\]
Repair then performs an Isabelle check on $P'$ (as a full theory). If verification succeeds, $P'$ is returned as a verified repair. If verification fails, repair records $\widehat{B}'$ into $\mathcal{M}$ (bounded by a maximum list length) and continues, provided $\mathrm{left}()$ is positive.

If a candidate changes the script but fails verification, repair may still return it as \emph{partial progress} with a diagnostic tag (e.g.\ \texttt{stage=1 partial-progress}), allowing the outer loop to apply $\mathsf{OpenMinimalSorries}$ and continue filling with explicit holes.

\subsubsection{Outer control: stage caps, focus management, and escalation}

The driver maintains per-hole attempt counts for each start stage. Let $\mathrm{tries}[(\mathrm{hid},s)]$ be the number of verified-gate failures (including no-change outcomes) at stage $s$ for hole $\mathrm{hid}$. Two caps $c_1$ and $c_2$ are enforced:
\[
c_1 = 2,\qquad c_2 = 3.
\]
Escalation proceeds as:
\[
\mathrm{stage}[\mathrm{hid}] \leftarrow
\begin{cases}
2 & \text{if } \mathrm{stage}=1 \text{ and } \mathrm{tries}[(\mathrm{hid},1)] \ge c_1,\\
2\ \text{and trigger whole regeneration} & \text{if } \mathrm{stage}=2 \text{ and } \mathrm{tries}[(\mathrm{hid},2)] \ge c_2.
\end{cases}
\]

\paragraph{Focus policy.}
Whenever partial progress opens new holes, the driver selects a nearest-hole heuristic around the previous location and sets $\mathrm{focus}$ to that hole’s fingerprint. This yields a coherent local search process rather than bouncing between unrelated holes.

\paragraph{Robustness to Isabelle instability.}
All Isabelle calls in fill and repair are wrapped with bounded timeouts. Exceptions (timeouts, value errors from malformed intermediate theories, or unexpected failures) trigger a controlled Isabelle restart and the corresponding attempt is counted as failed, without terminating the overall planning loop.

\subsubsection{Algorithmic summary}

The fill+repair loop can be summarized as a bounded search procedure over scripts:

\begin{quote}
\textbf{State:} current script $P$, hole set $\mathrm{Holes}(P)$, planner state $\Pi$. \\
\textbf{Actions:} \textsc{FillHole} via stepwise prover; \textsc{RepairBlock} via LLM; \textsc{OpenMinimalSorries}; \textsc{RegenerateWhole}. \\
\textbf{Transition:} apply an action to obtain $P'$; if \textsc{Verify}$(P')$ succeeds, commit; otherwise normalize by opening holes and update $\Pi$. \\
\textbf{Objective:} reach a script $P$ such that \textsc{Verify}$(P)=\text{true}$ and \texttt{sorry}$\notin P$ under a global time budget.
\end{quote}

This design mirrors the stepwise prover’s philosophy: the LLM is used to propose candidate edits, while Isabelle remains the sole judge of correctness. The main difference is the \emph{granularity} of proposals: the stepwise prover proposes single commands; the planner’s repair proposes structured Isar blocks, enabling larger ``jumps'' in the proof space when local command-level search stalls.

\section{Data Processing and Generation}
\label{sec:data}

Both the stepwise prover and the proof planner are instrumented to produce rich execution traces that serve simultaneously as debugging artifacts, benchmark results, and training data for learning-based components. This section describes the data model, logging pipeline, dataset generation procedures, and benchmarking infrastructure implemented across the \texttt{prover/} and \texttt{planner/} modules.

\subsection{Design principles}

The data pipeline is guided by four design principles:

\begin{enumerate}
  \item \textbf{Verifier-grounded data}: all logged signals are derived from concrete Isabelle executions rather than model-internal confidence.
  \item \textbf{Fine-grained supervision}: every attempted action (tactic, block edit, outline) is logged, not only successful proofs.
  \item \textbf{Phase separation}: data from tactic-level proving, planning, filling, and repair are distinguishable but share a common schema.
  \item \textbf{Reproducibility}: logs are append-only, version-agnostic, and sufficient to replay or re-evaluate decisions offline.
\end{enumerate}

As a result, the system naturally produces datasets suitable for supervised learning, offline reinforcement learning, ablation studies, and longitudinal benchmarking.

\subsection{Run-level logging}

Each invocation of the prover or planner produces a \emph{run record}, corresponding to a single goal attempt. Let $G$ be a goal statement. A run record $R(G)$ contains:

\begin{itemize}
  \item goal identifier and textual goal,
  \item prover/planner mode and configuration (beam size, depth, temperatures, budgets),
  \item model identifiers for LLM, reranker, and premise selector,
  \item wall-clock runtime and timeout status,
  \item success flag and final proof text if successful,
  \item summary statistics (depth reached, number of expansions, number of repairs, number of regenerations).
\end{itemize}

Run records are serialized as one JSON object per line (JSONL), enabling streaming writes and post-hoc aggregation without loading entire logs into memory.

\subsection{Attempt-level logging}

More granular supervision is captured at the \emph{attempt level}. An attempt corresponds to a single proposed action evaluated by Isabelle. The precise meaning of an attempt depends on context:

\begin{itemize}
  \item in the stepwise prover: applying a single candidate command to a proof state;
  \item in the planner fill phase: inserting a candidate proof fragment into a hole;
  \item in the planner repair phase: replacing a structured block or subproof.
\end{itemize}

Each attempt record includes:
\begin{itemize}
  \item the goal $G$ and current script prefix or outline,
  \item the proposed action text,
  \item attempt type (step, finisher, fill, repair, regeneration),
  \item success flag under Isabelle verification,
  \item number of subgoals before and after the attempt (when applicable),
  \item elapsed Isabelle execution time,
  \item cache hit indicators,
  \item current search depth or repair stage,
  \item auxiliary features (retrieval scores, reranker scores, hint usage).
\end{itemize}

Formally, each attempt yields a labeled tuple
\[
(x, y, \Delta),
\]
where $x$ is the feature representation of the attempt, $y \in \{0,1\}$ is the verifier outcome, and $\Delta$ captures state change metrics such as subgoal reduction.

These attempt logs are the primary source of training data for tactic rerankers, premise encoders, and repair heuristics.

\subsection{Dataset generation for continual learning}

The logged attempts are post-processed into structured datasets tailored to different learning tasks.

\paragraph{Reranker datasets.}
For tactic and block reranking, each attempt produces a supervised example $(x, y)$, where $x$ is the numeric feature vector and $y$ indicates verifier acceptance. Depending on the training objective, labels may be transformed into:
\begin{itemize}
  \item binary classification targets,
  \item regression targets approximating $Q$-values,
  \item advantage-weighted targets for offline policy learning.
\end{itemize}

Episodes corresponding to single proof attempts can also be reconstructed, yielding trajectories
\[
(s_0,a_0,r_0,s_1),\dots,(s_T,a_T,r_T,s_{T+1}),
\]
where rewards $r_t$ are derived from subgoal reduction or proof completion.

\paragraph{Premise selection datasets.}
For premise selection, successful attempts provide weak supervision: any lemma explicitly referenced in a successful step is treated as a positive premise for the corresponding goal or state. Negative premises are sampled from the retrieved pool but unused in the proof. This yields contrastive pairs $(g,p^+)$ and $(g,p^-)$ for training bi-encoders and cross-encoders.

\paragraph{Planner repair datasets.}
Planner logs additionally encode:
\begin{itemize}
  \item block types (have/show, case, subproof),
  \item effective goals for holes,
  \item counterexample diagnostics,
  \item failure histories and banned candidates.
\end{itemize}

These signals support future work on learning block-level repair policies or regeneration strategies.

% \subsection{Benchmark harness}

% The project includes a unified benchmark harness that evaluates the prover and planner on curated goal suites. A benchmark suite is a text file
% \[
% \mathcal{B} = \{G_1,\dots,G_n\},
% \]
% where each $G_i$ is a standalone Isabelle goal.

% For each $G_i$, the harness:
% \begin{enumerate}
%   \item initializes a fresh run with fixed configuration,
%   \item executes the prover or planner under a timeout,
%   \item records run-level and attempt-level logs,
%   \item optionally verifies and minimizes successful proofs.
% \end{enumerate}

% Benchmarks support repeated runs with different random seeds, temperatures, or models, enabling controlled comparisons and ablation studies.

% \subsection{Regression testing and reproducibility}

% To guard against performance regressions, the harness supports regression mode, in which previously solved goals are re-run under new code or model versions. Differences in success rate, runtime, and proof structure can be detected automatically.

% All logs include explicit configuration snapshots, allowing a run to be reproduced by replaying the same goal with the recorded parameters. This design makes experimental results auditable and facilitates long-term empirical studies.

% \subsection{Data flow summary}

% The overall data flow can be summarized as:
% \[
% \text{Isabelle execution}
% \;\rightarrow\;
% \text{attempt logs}
% \;\rightarrow\;
% \text{run summaries}
% \;\rightarrow\;
% \text{derived datasets}
% \;\rightarrow\;
% \text{trained models}
% \;\rightarrow\;
% \text{improved prover/planner}.
% \]

By treating proof search as a data-generating process rather than a black-box solver, the system enables continual improvement through learning while preserving the soundness guarantees of formal verification.

\section{Integration with the Isabelle/jEdit UI}
\label{sec:isabelle-ui}

To support interactive use, the system includes a lightweight integration with the Isabelle/jEdit user interface. This integration allows users to invoke the stepwise prover and the proof planner directly from the editor, at the location of the current lemma, and to insert the generated proof text back into the buffer. The design deliberately keeps the UI layer thin: all reasoning and verification happen in the existing prover and planner components, while the UI acts purely as a convenience interface.

\subsection{Overall architecture}

The integration consists of two parts: a local HTTP server and a set of jEdit macros. The HTTP server runs as a long-lived process and exposes a small API that wraps the prover and planner entry points. It also maintains a persistent Isabelle server session, so that repeated UI invocations reuse the same HOL session and avoid repeated startup overhead. From the UI’s perspective, this server is the single point of contact for all proof automation requests.

On the editor side, Isabelle/jEdit is extended with BeanShell macros. These macros extract the goal of the lemma near the caret, send it to the local server, and paste the returned proof text back into the editor. Communication is local-only and uses simple JSON payloads over HTTP.

\subsection{Server-side functionality}

The UI server is implemented using a small FastAPI application. On startup, it initializes a persistent Isabelle server and opens a HOL session that is reused for all subsequent requests. The server exposes two main endpoints.

The first endpoint provides access to the stepwise prover. It accepts a goal string together with basic search parameters such as time budget and model selection, invokes the prover, and returns the resulting sequence of proof commands. Before returning the result, the server filters the output to retain only lines that are directly usable inside an Isabelle proof, typically commands beginning with \texttt{apply} or \texttt{by}. This ensures that pasted output is immediately syntactically valid.

The second endpoint provides access to the proof planner. Depending on the request mode, it either returns a proof outline or attempts a full plan-and-fill run. The endpoint forwards planner-specific parameters such as outline diversity, repair options, and micro-RAG configuration, while falling back to server-side defaults when parameters are omitted. The response contains the current best proof script produced by the planner, whether or not all gaps have been filled.

The server is designed to be robust against Isabelle instability. Timeouts and unexpected failures are handled by restarting the Isabelle session when necessary, without crashing the server process itself.

\subsection{Editor macros}

The jEdit macros implement a uniform editor-side workflow. When a macro is invoked, it scans upward from the caret to locate the nearest \texttt{lemma} or \texttt{theorem} declaration and extracts the quoted goal string. It then determines whether the caret lies inside the corresponding proof block or outside it. This decision controls where the generated text will be inserted: either at the caret position or immediately after the lemma header.

Three macros are provided. One macro invokes the stepwise prover and inserts the suggested proof commands. A second macro invokes the planner in outline-only mode and inserts a structured proof skeleton. The third macro invokes the planner in full plan-and-fill mode and inserts the resulting script, which may already be complete or may still contain explicit gaps.

Before insertion, planner outputs are lightly normalized on the editor side. In particular, redundant lemma headers are stripped, and only the \texttt{proof}–\texttt{qed} block is inserted. Indentation is adjusted to match the surrounding context in the editor buffer.

\subsection{Intended usage}

The UI integration is intended as an interactive aid rather than a replacement for batch evaluation. A typical workflow is to write or import a lemma statement in Isabelle/jEdit, place the caret inside the lemma, and invoke one of the macros. The prover or planner then attempts to construct a proof, and the resulting text is inserted directly into the editor, where it can be inspected, edited, or re-run.

By keeping the UI layer minimal and delegating all substantive reasoning to the underlying system, the integration provides a smooth interactive experience without compromising soundness or duplicating logic already present in the prover and planner.

\begin{figure}
    \centering
    \includegraphics[width=0.9\linewidth]{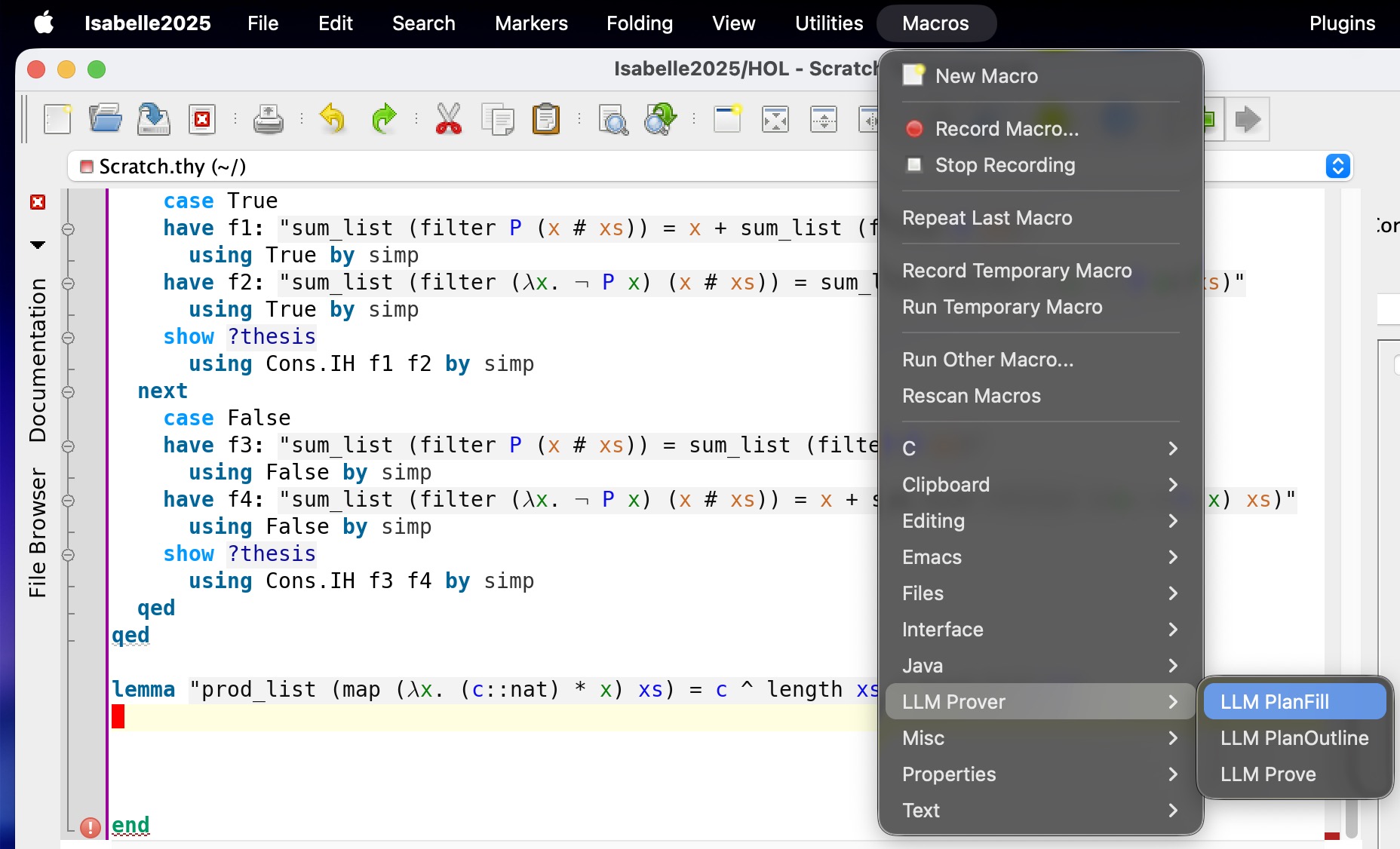}
    \caption{An illustration of the integration with Isabelle/jEdit.}
    \label{fig:PlanFill}
\end{figure}

\begin{figure}
    \centering
    \includegraphics[width=0.9\linewidth]{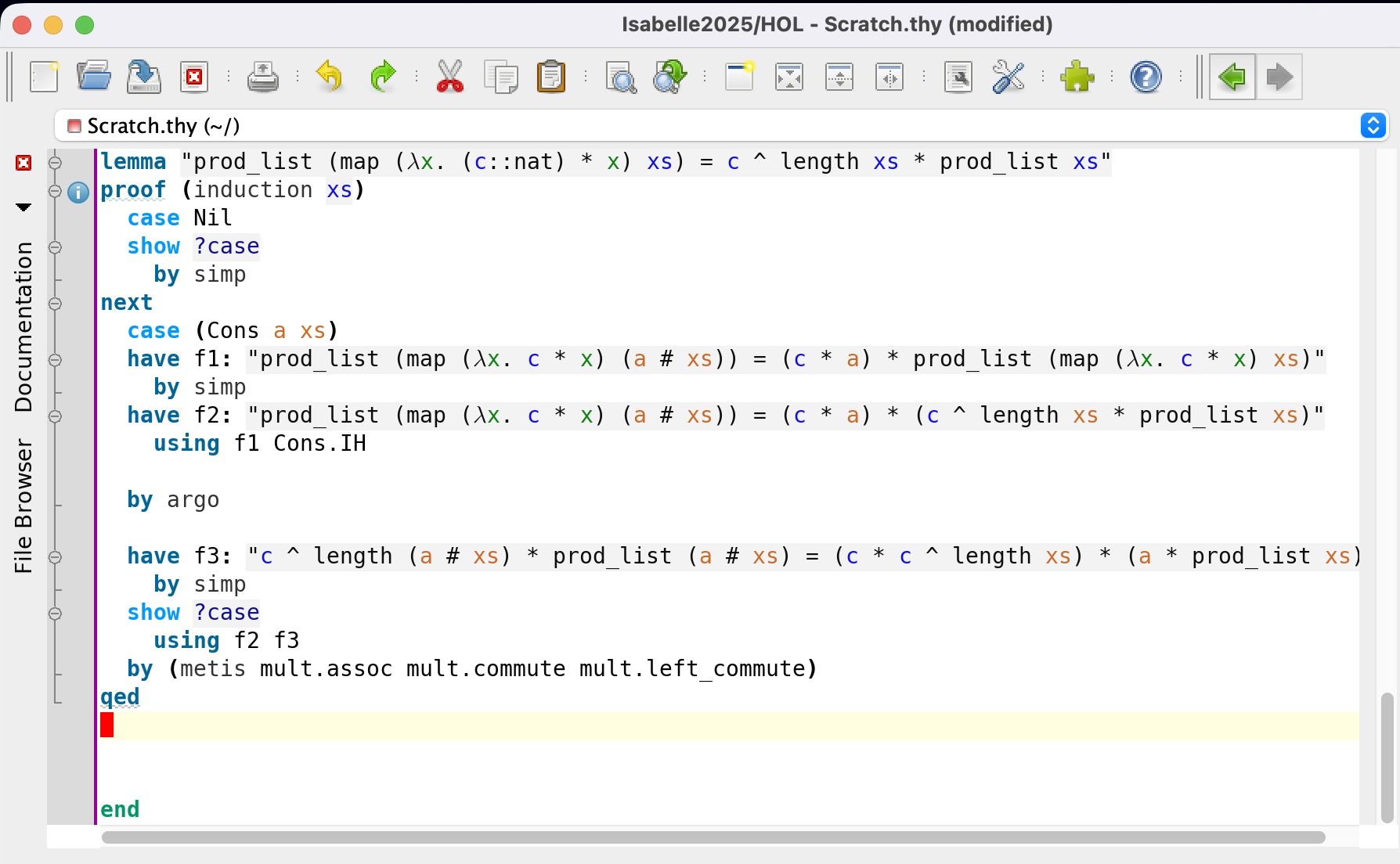}
    \caption{An example of a proof inserted by the LLM prover.}
    \label{fig:result}
\end{figure}

As as example, assuming that the user has started the local server as described above, the user may move the cursor to a proof state and click LLM $\rightarrow$ PlanFill from the Macro menu, as shown in Figure~\ref{fig:PlanFill}. When the computation is complete, the LLM prover will insert a proof, if it can find a valid one, at the correct position of the document, as shown in Figure~\ref{fig:result}.

\section{Conclusion}
\label{sec:conclusion}

This work presented an LLM-powered theorem proving system for Isabelle/HOL that combines verifier-guided stepwise search with a higher-level proof planner based on Isar outlines, gap filling, and repair. The system is designed to operate fully automatically: given a goal, it attempts to synthesize a complete, kernel-checked proof without user interaction. Throughout the design, Isabelle remains the sole authority on correctness, while large language models are used strictly as proposal mechanisms whose outputs are filtered, validated, and repaired under tight syntactic and semantic constraints.

A key contribution of the project is demonstrating that LLM-driven proof search can already surpass classical automation in specific regimes. In particular, the stepwise prover is able to solve certain higher-order goals that are beyond the reach of Sledgehammer, even when the latter is given generous timeouts. Figure~\ref{fig:result} illustrates a representative example: a lemma that cannot be proved by Sledgehammer but can be proved by the LLM prover through a sequence of semantically informed proof steps. This shows that LLM-based guidance is not merely a convenience layer over existing automation, but can explore proof strategies that differ qualitatively from those encoded in traditional heuristics.

At the same time, the project exposes clear and fundamental limitations. Most notably, the fill-and-repair component of the proof planner remains far less effective than originally intended. Despite careful engineering of effective-goal extraction, staged repair, CEGIS-style loops, and whole-proof regeneration, even state-of-the-art LLMs such as GPT~5.2~Extended Thinking and Gemini~3~Pro struggle to reliably implement the design the fill and repair features described in Section~\ref{subsec:cegis}. In practice, these models often fail to generate code with such complex algorithmic design, especially when the code base reaches more than 20 files and the new code needs to respect the existing APIs. As a result, the planner’s repair stages rarely succeed beyond trivial cases, and most successful proofs are obtained either directly from the initial outline or from the stepwise prover operating without planner assistance.

These limitations are not merely implementation artefacts but reflect deeper challenges. Repairing an Isar proof requires simultaneously reasoning about global structure, local proof states, and Isabelle’s context-sensitive proof modes, all under a strictly typed and indentation-sensitive syntax. Current LLMs, even with long contexts and advanced reasoning capabilities, appear to lack a sufficiently precise internal model of these constraints to make repair reliable at scale. This suggests that improving fill-and-repair will likely require new forms of training data, tighter symbolic abstractions, or deeper integration between the planner and the stepwise prover, rather than simply stronger language models.

Despite these shortcomings, the current implementation is already a useful tool for Isabelle/HOL users. The stepwise prover alone can discharge goals that defeat existing automation, and the planner provides a foundation for future work on structured proof synthesis. More broadly, the project demonstrates that treating theorem proving as a data-generating, verifier-in-the-loop process enables systematic experimentation and incremental improvement, even when some components fall short of their ultimate goals. We view this work as a step toward accessible, fully automatic theorem proving in expressive logics, and as a concrete case study of both the promise and the current limits of LLM-based reasoning in formal verification.

\bibliographystyle{plainnat}
\bibliography{references}

\end{document}